# Learning the Pseudoinverse Solution to Network Weights

Authors: J. Tapson and A. van Schaik

Affiliation: Bioelectronics and Neuroscience Group
MARCS Institute
University of Western Sydney
Building XB, Kingswood Campus
Kingswood 2751 NSW
Australia

Corresponding Author: J. Tapson

Email Address: j.tapson@uws.edu.au

Phone: +61 2 4736 0238

Fax: +61 2 4736 0833

**Abstract**

The last decade has seen the parallel emergence in computational neuroscience and machine learning of neural network structures which spread the input signal randomly to a higher dimensional space; perform a nonlinear activation; and then solve for a regression or classification output by means of a mathematical pseudoinverse operation. In the field of neuromorphic engineering, these methods are increasingly popular for synthesizing biologically plausible neural networks, but the "learning method" - computation of the pseudoinverse by singular value decomposition - is problematic both for biological plausibility and because it is not an online or an adaptive method. We present an online or incremental method of computing the pseudoinverse, which we argue is biologically plausible as a learning method, and which can be made adaptable for non-stationary data streams. The method is significantly more memory-efficient than the conventional computation of pseudoinverses by singular value decomposition.

## 1. Introduction

One of the core problems for neural network practitioners, in both the neuromorphic and machine learning fields, is the following: given a known input-output relationship, or input and output data sets, how do we synthesize a network that will optimally model the relationship? This synthesis problem is generally divided into two parts – what the architecture or structure of the network should be, and how we should establish the weights for connections between the neurons.

In the last decade we have seen the independent and parallel emergence, in the neuroscientific and machine learning fields, of similar architectures and weight-learning algorithms that neatly and efficiently solve both of these synthesis problems, for a large class of relationships. The methodologies, named by their discoverers *Neural Engineering* (Eliasmith & Anderson, 2003) in computational neuroscience and the *Extreme Learning Machine* (ELM) (Huang, Zhu & Siew, 2006) in the machine learning field, both synthesize three-layer feedforward networks which are superficially similar to the classic multilayer perceptron - with input, hidden and output layers (the neural engineering approach also makes use of recurrent connections in the hidden layer). What makes these architectures unique is that the input layer signals are connected to an unusually large number of hidden layer neurons, using randomly initialized connection weights. This has the effect of randomly spreading or projecting the inputs from their original input dimensionality to a hidden layer of very much higher dimensionality. It is then possible to find a hyperplane in the higher dimensional space which approximates a desired function regression solution, or represents a classification boundary for the input-output relationship. The output neurons need therefore compute only a linearly weighted sum of the hidden layer values in order to solve the problem. These linear weights can then be determined analytically by computing the product of the pseudoinverse of the hidden layer values with the desired output values. The methodology can be summarized very simply as follows:

1. Connect the input layer of neurons to a much larger hidden layer, using random weights.
2. Analytically solve the output weights (between the large hidden layer and linear output neurons) by calculating the pseudoinverse of the product of the hidden layer activations and the target outputs.

The method works for both continuous-valued and spiking neural networks, where the spiking signals are rate encodings of an underlying variable (Eliasmith & Anderson, 2003). It has the advantage of requiring no learning as such – the full and final solution is obtained with one analytical calculation step.

We will refer broadly to this class of methods as linear solutions to higher dimensional interlayer networks (LSHDI). The ELM form of the method has been rapidly and widely adopted in the machine learning field because it offers, in a single forward computational step, a least-squares optimal, non-parameterized solution to the problem of learning network weights; and it is computationally equivalent (but much faster to train) than the widely used support vector machine. It is therefore very quick to compute, and avoids the problems of stability and convergence on local minima which plague the users of error backpropagation methods, for example.

In bio-inspired and neuromorphic networks, the LSHDI method solves a significant modelling problem, which is that it has been extremely difficult to construct simulated biological neural networks to model specific relationships – while the neuronal elements of these networks are well defined, there has been no widely applicable method to synthesize a network to solve a given problem. In one form (the *Neural Engineering Framework*, or NEF) the method is emerging as the core of a generic compiler for silicon neural systems (Choudhary et al., 2012; Galluppi et al., 2012). These silicon neural systems have reached the point where they are not limited by the number or complexity of the neurons or interconnects on the hardware platform, but by the ease with which they can be programmed. The NEF approach has been enthusiastically adopted by neuromorphic engineers working on silicon-based neural networks.

Significantly, we are starting to see some evidence from neurophysiology that structures embodying the LSHDI principle may exist in the brain. For example, recent work by Rigotti *et al*. (2010) in modeling recorded cortical activity in monkeys performing context-sensitive tasks, shows that complex rule-based tasks require both sensory stimuli and internal representation of states; and that a significant number of random connections placed between input sources and a hidden interlayer, and random recurrent connections between interlayer neurons, are necessary for optimal performance. They describe these interlayer neurons as generating *mixed selectivity*, which is equivalent to increasing the dimensionality of the state representation.

An obstruction to acceptance of the LSHDI method as being biologically plausible, is the necessity to compute the pseudoinverse of a matrix in order to synthesize the network. The Moore-Penrose pseudoinverse is a relatively recent addition to linear algebra (Penrose, 1956), and is usually computed using singular value decomposition (SVD). It seems intrinsically unlikely that we would find in the cortex a plausible equivalent to this mathematically complex process.

In this paper we demonstrate an LSHDI algorithm which incrementally learns the pseudoinverse solution to the weights problem, in the context of online presentation of new training input, and we show that it is plausible as a physiological process in real neurons. The solution to the network weights is exactly the same as that which would be calculated by singular value decomposition. It converges with a single forward iteration per input data sample, and as such is ideal for real-time online computation of the pseudoinverse solution. It requires significantly less memory than the SVD method, as its memory requirement scales as the square of the size of the hidden layer, whereas the SVD memory requirement scales with the product of the hidden layer size and the size of the training data set. Given that the data set size should significantly exceed the hidden layer size, the former is advantageous. We call the algorithm the OLP (online pseudoinverse) method.

The OLP algorithm is adapted from an iterative method for computing the pseudoinverse, known as Greville's method (Greville, 1960). The existence of the OLP algorithm, and its biological plausibility, suggests that there is no reason why a biological neural network would not arrive at the same weights that are computed using a singular value decompostion, and therefore that this method of synthesizing network structures may be used without the fear that it is biologically implausible.

In addition, we show for the first time that a single modification to this algorithm allows it to adapt over time to changes in the modeled relationship. Given that in the real world, input-output relationships are seldom stationary processes, an adaptive network is considerably more robust (and plausible) than one that finds a static solution.

Pseudoinverse methods were widely used in an earlier era of neural network research, to the extent that a significant class of Kohonen-type linear associative memories were known as pseudoinverse neural networks (PINNs) (Kohonen, 1988 and 2000). The key variations in the LSHDI use of the method, are in the initial spreading to higher dimension, and the associated nonlinear activation performed on the higher dimension signals. We are not aware of any prior recasting of the pseudoinverse method into a biologically plausible framework.

We begin by describing the OLP method, demonstrate it in practice, and then motivate its biological plausibility.

## 2. Mathematical Foundations

The structure of LSHDI networks can be generalized from that of a standard multilayer perceptron – see for example [9]. If we consider a particular sample of input data to be $x_{(t)}$ where $t$ is a time or series index, then forward propagation of the local signals through the network can be described by:

$$y_{n,(t)} = \sum_{j=1}^{M} w_{nj}^{(2)} g\left(\sum_{i=1}^{d} w_{ji}^{(1)} x_{i,(t)}\right) \quad (1)$$

where $y_{(t)}$ is the output layer vector corresponding to input $x_{(t)}$, and each element $y_n$ is a linear sum of the $M$ hidden layer outputs weighted by $w_{nj}$. $n$ is the output vector index, $j$ the hidden layer index, and $i$ the input vector index. The hidden layer outputs depend on the neuron's activation function $g()$ and the randomly determined interconnecting weights $w_{ji}$ between input and hidden layer. The superscripts indicate the layer. In this representation all the weights are static. The number of hidden layer neurons, *M*, is deliberately chosen to be large compared to the size of the input layer, which is denoted *d*; for example, *M* > 10*d* is not unusual.

The random selection of static input weights $w_{ji}$ reduces the network training requirement to be the optimization of the output weights $w_{nj}$, and the linear output summation reduces this problem to one of linear regression. This is why the ELM has found such favour with practitioners; training consists of a single parameter-less analytical calculation, rather than some type of gradient descent with backpropagation, or a stochastic learning algorithm.

The weight optimization problem can be stated as follows: given a series of hidden layer outputs

$$a_{j,(t)} = g\left(\sum_{i=1}^{d} w_{ji}^{(1)} x_{i,(t)}\right) \qquad (2)$$

we can form a matrix $A = [a_{(0)} \; a_{(1)} \ldots a_{(k)}]'$ in which each row contains the output of the hidden layer at one instant in the series, with the last row containing the most recent instant in the series; and then given a similar matrix *Y* consisting of the corresponding output values $Y = [y_{(0)} \; y_{(1)} \ldots y_{(k)}]'$: what set of weights $w_{nj}$ will minimize the error in:

$$Aw = Y \; ? \qquad (3)$$

This may be solved analytically by taking the Moore-Penrose pseudoinverse $A^+$ of *A*:

$$w = YA^+. \qquad (4)$$

When *A* and *Y* are explicitly and exhaustively known (static) data sets, then a singular value decomposition suffices to determine $A^+$. In this paper, we show that $A^+$ may be learnt incrementally in a real-world or real-time application as new data becomes available; and with a modification, it may be made adaptable for non-stationary data sets.

There are a number of methods for incremental calculation of the pseudoinverse solution to this problem (Greville, 1960; Kohonen, 2000) and we are aware of an existing method by Huang *et al.* (2007) for incremental solution of the ELM. (We note however that current online or incremental methods focus on learning through the addition of further neurons, whereas this research focuses on learning by online recalculation of network weights.) We have selected one incremental solution method, Greville's algorithm, for use in this work.

Greville's original method for calculating the pseudoinverse (Greville, 1960) is well known, and has been used by several groups to synthesize neural networks in which extra hidden layer neurons are added incrementally. The standard form of Greville's algorithm does not lend itself to applications in which the network structure is static; but a variation presented by Ben-Israel and Greville (2003) may be adapted for the purpose, as follows. Given the input and output data streams *a* and *y* respectively for some process sampled *k-1* times, when the next set of data $a_k$, $y_k$ becomes available, we form two vectors partitioned as follows (for real-valued data):

$$A_{(k)} = \begin{bmatrix} A_{(k-1)} \\ a_k^T \end{bmatrix} \qquad (5)$$

$$Y_{(k)} = \begin{bmatrix} Y_{(k-1)} \\ y_k^T \end{bmatrix}. \tag{6}$$

We want to calculate $\widehat{w}_k$ given $\widehat{w}_{k-1}$ where

$$\widehat{w}_k = Y_{(k)} A_{(k)}^+, \quad \widehat{w}_{k-1} = Y_{(k-1)} A_{(k-1)}^+. \tag{7}$$

We use a weight update algorithm:

$$\widehat{w}_k = \widehat{w}_{k-1} + \left(y_k - \widehat{w}_{(k-1)} a_k\right) b_k \tag{8}$$

where

$$b_k = \frac{a_k^T \theta_{(k-1)}^T}{1 + a_k^T \theta_{(k-1)} a_k}. \tag{9}$$

The term $\theta$ here serves the purpose of the pseudoinverse $A^+$; we have named it the *inhibition matrix* as it has the effect of distributing activation energy independently and thereby producing lateral inhibition in the output weights. It is initialized as $\theta_{(0)} = I/M$ where $I$ is the identity matrix.

The inhibition matrix is updated at each timestep:

$$\theta_{(k)} = \theta_{(k-1)} - (\theta_{(k-1)} a_k) b_k. \tag{10}$$

We have adapted this algorithm for non-stationary data as follows: a decay or "forgetting" term is built into the inhibition matrix update;

$$\theta_{(k)} = \theta_{(k-1)} - (\theta_{(k-1)} a_k) b_k + I \left(\frac{1 - e^{-|E|}}{M}\right) \tag{11}$$

Where $E$ is the output error, normalized to activation levels:

$$E = \frac{y_k - \widehat{w}_{(k-1)} a_k}{(1 + a_k^T \theta_{(k-1)} a_k)^{\frac{1}{2}}}. \tag{12}$$

The effect of this term is that when errors are high, suggesting that the model is no longer accurate, the inhibition matrix is pushed towards its initial value (and lateral inhibition is reduced), suggesting that not all the activation energy is appropriately distributed. It has the consequence that more recent inputs have a higher effect on the weights than earlier inputs, as in the general class of infinite-impulse response (IIR) filters. Similarly, it shares with IIR filters the potential for instability in recursion, although we have found this to be very uncommon.

Obviously, this adaptive solution is no longer a mathematically exact pseudoinverse solution to equation (3) for the complete data sets *A* and *Y*. It is, however, much more useful for non-stationary real-world processes, as will be seen in the examples.

## 3. Applications of the Method: Handwritten Digits

Recognition or classification of handwritten digits is a standard machine learning problem, and in the form of the MNIST database (LeCun & Cortes, 2012) it has become a benchmark problem. For the LSHDI methods, this problem represents a massive computational burden: the digits are 28 x 28 pixels, and the standard learning set contains 60000 sample digits. Given a standard rule-of-thumb "fan-out" of 10 hidden-layer neurons per input, this would require a network of 784 input neurons, 7840 hidden-layer neurons, and the pseudoinversion of a 7840 x 60000 matrix. At ~5 x $10^9$ elements, this significantly exceeds the size of matrix that can be inverted without incurring significant memory issues on most computational platforms (particularly given that the SVD method requires solution of three matrices of approximately this size); in the absence of several hundred Gb of RAM it would require extensive disk read/write operations which would impact severely on computation time.

By using the OLP method, we can perform the LSHDI classification of this data set in RAM on a standard notebook computer (data presented in this paper were generated on a MacBook Air). The image size can be reduced to 24 x 24 pixels without significant loss of information, giving a network structure of 576 input neurons, 5760 hidden-layer neurons and 10 outputs. This reduces the maximum matrix size requirement to 5760 x 5760 for the OLP method, as opposed to 5760 x 60000 (x 3) for the SVD solution of the pseudoinverse.

Using the structure shown in Fig. 1, the typical test error on the MNIST data set is 2.75%, which compares favorably with most results from 2-layer feedforward networks on this problem (LeCun et al. 1998). Given that the structure has been applied absolutely naïvely to this problem – in fact, in the same way as it would be applied to any generic problem with similar input-output pairs – the result reinforces the usefulness of this technique. We note that apart from the number of hidden-layer neurons, there are no parameters to be tuned and there is no learning process except for the single pseudoinversion operation. The nonlinear hidden-layer neurons used the standard sigmoidal activation function.

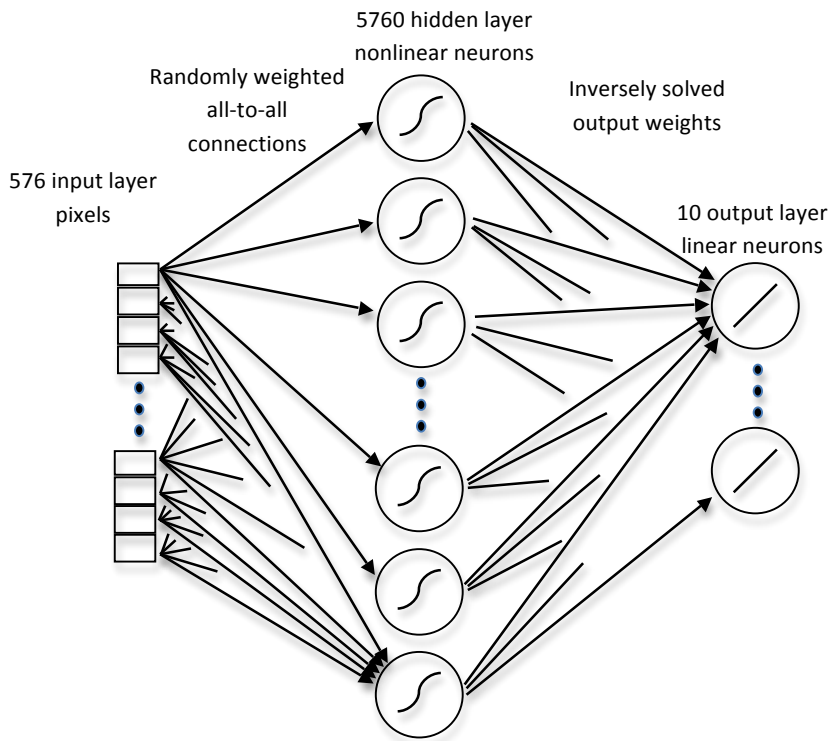

**Fig. 1:** Structure of an LSHDI network as used to classify the MNIST handwritten digit database. The inputs are the pixels; they are connected to a higher-dimensional interlayer of nonlinear neurons, using randomly weighted connections. The output layer consists of linear neurons and the output layer weights are solved analytically using the pseudoinverse operation.

The progression of online training is also illustrated in the MNIST database example. We can present the training data one at a time, and update the linear solution with each presentation. The progressive improvement for this method is illustrated in Fig 2. This shows networks of three sizes, in which training data consisting of single digits randomly selected from the test set, were presented one at a time and the new solution computed after each new input.

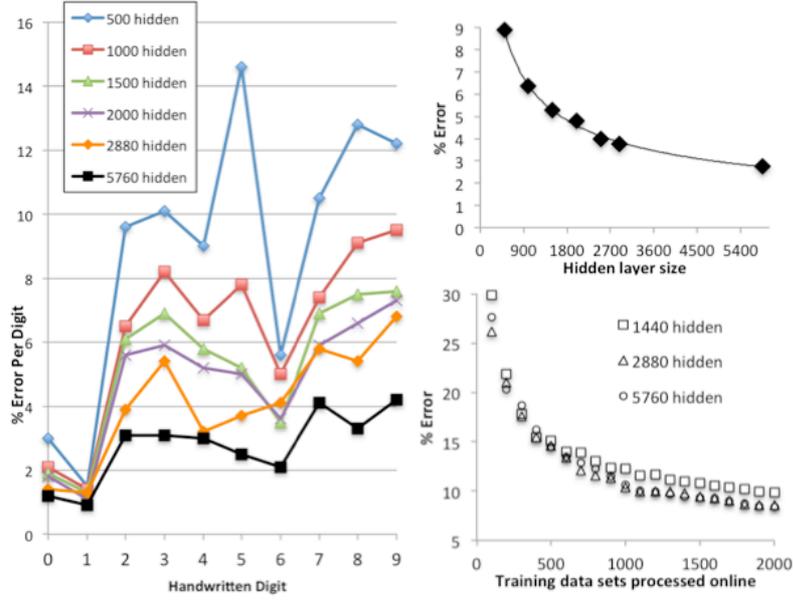

**Fig. 2:** Performance of the LSHDI network in classifying the MNIST handwritten digit benchmark problem. The left hand graph shows improving performance per digit, as the number of hidden layer neurons increases. Simple forms such as 0 and 1 are more accurately classified than complex forms such as 3 and 5. The RH upper figure shows overall improvement in accuracy with hidden layer size $M$; the errors decrease in proportion to $M^{-0.5}$. The RH lower figure shows the online learning in progress, for three network sizes; it can be seen that after just 2000 randomly selected digits from the training set, the test set is classified with 90% accuracy by networks with large hidden layer size.

## 4. Applications of the Method: Predicting a Chaotic Time Series

We demonstrate the application of the adaptive method with a simple and well-known time series prediction problem, which is to predict the output of the Mackey-Glass (MG) time series (Mackey & Glass, 1977). The MG time series is generated by a nonlinear delay differential equation, shown below, which displays chaotic oscillations:

$$\frac{dx(t)}{dt} = \frac{ax(t-\tau)}{1+x(t-\tau)^{10}} - bx(t) \qquad (13)$$

This is typically solved using a 4$^{th}$ order Runge-Kutta method. Standard parameters, used by other research groups, are $a = 0.2$; $b = 0.1$; $\tau = 17$; and a computational timestep of 0.1 arbitrary intervals is used.

We have used the LSHDI method to model the series in the usual fashion, which is to tap the series at a fixed set of delays, and then predict the value of the series at some interval in the future. Once again, we have used a standard set of taps and delays in order to produce results comparable with other methods. The data in Figure 3 were modeled with a network using four taps in the range 100 to 1 lagging time intervals, and a hidden layer of 100 neurons. The task was to predict the value ten intervals ahead.

Figure 3 shows the MG time series and the corresponding predictions for the LSHDI method, in the stationary and the adaptive OLP forms. It can be seen that the adaptive method produces a significantly better prediction. It should be noted that this chaotic time series is not in any real sense being modeled globally by the adaptive method, as it is by the static method; the adaptive LSHDI method is finding a local solution to the problem and "forgetting" prior data with an exponential decay of memory.

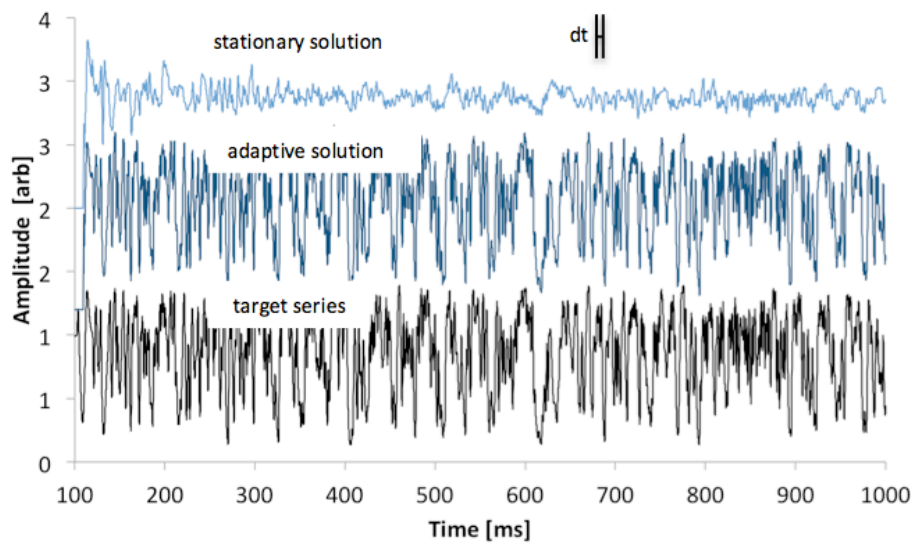

**Fig. 3:** The Mackey-Glass chaotic time series predicted with stationary and adaptive LSHDI solutions; the magnitude of the forward prediction d$t$ is indicated by the vertical bars at top. It can be seen that by focusing on an adaptive local prediction rather than global model, the adaptive method is significantly more accurate.

We note that none of the examples above use spiking neurons. Eliasmith and colleagues (2003) have used the LSHDI method extensively with rate-encoded spiking neurons, without any apparent loss of utility or accuracy; the weights in these networks are solved using rate variables rather than actual spike trains. This begs the question whether there is any point in using rate-encoded spikes rather than the underlying rate variable; in practice, there may be dynamical effects such as synchrony which occur when coding is performed with spikes, which will not occur directly with the underlying variable.

## 5. Biological Plausibility

The question as to whether a given computational neural process is biologically plausible may seem ill-posed, but it has been addressed in some detail by O'Reilly (1998). He defines biological realism as an overarching principle, and then identifies five architectural principles which can provide converging evidence of biological realism. We list the architectural principles below, and comment on the adaptive OLP method in that context.

1. **Distributed representation of information**

    The general LSHDI model gives an excellent framework for distributed representation of information, in that input information is spread into a higher dimensional space, and then reduced in dimensionality by construction from that space. Each input variable is represented to some extent in the output of every hidden layer neuron, and loss of a single hidden layer neuron would not significantly degrade the reconstruction of the input variables (the architecture is, after all, a type of associative memory). While it is not a core issue in this research, the architecture convincingly embodies the principle of distributed representation of information.

2. **Inhibitory competition between units**

    The OLP algorithm uses the inhibition matrix to distribute the hidden layer activation energy competitively between weights (the competition being driven by the divisive normalization of activation energy, so that the total energy available is limited). This decomposition by divisive normalization has been identified and described by Schwarz and Simoncelli (2001) in sensory neural systems, amongst others. It has the effect of creating a soft winner-take-all effect, and in the context of Boltzmann machines has been suggested to act as a kind of principle component analysis (Kohonen, 2000).

3. **Bidirectional activation propagation**

    While the network itself is feedforward only, the inhibition matrix represents a modulation of the output gains of the hidden layer, for the purposes of learning the output layer weights, and can be recast as a backpropagation of error. It should be noted that the backpropagation of error in this model is only local, i.e., the error is used to adjust only the weights immediately prior to the output layer. This avoids the biological-plausibility weakness of the standard backpropagation algorithm, in which output error is backpropagated through several layers in a biologically implausible fashion.

4. **Error-driven task learning (supervised learning)**

    The weight update step in the OLP algorithm presented here is a very direct form of error-driven weight modification – it can be summarized as:
    $$\Delta w = E/N$$
    where $\Delta w$ is the weight change, $E$ is the error and $N$ a normalization factor. It is particularly suitable that the weight update is local and does not depend on reverse propagation of an error through multiple layers, as is required in conventional backpropagation.

5. **Hebbian learning (unsupervised learning)**

> The soft winner-take-all effect of the inhibition matrix has the effect of encouraging strong weight development between strongly firing hidden layer neurons and the output layer, which is a straightforward form of Hebbian learning.

This suggests to us that the incremental learning of the pseudoinverse meets O'Reilly's criteria to a reasonable degree, and that it can be re-cast as a combination of processes which are recognized to occur in biological neural systems.

## 6. Conclusions

The incremental learning method presented here has two significant uses. The first is largely symbolic; by presenting a biologically-plausible learning method which derives pseudoinverse solutions, we have opened the door to the use of the associated LSHDI methods in biological models, in such a way that awkward questions about the physiological likelihood of the pseudoinverse computation can be addressed with some confidence. The second use of this method is practical; by presenting a new algorithm for online learning and adaptive updating of the pseudoinverse solution in the face of non-stationary processes, we have enabled its use with greater simplicity and accuracy in real-world situations. In addition, the method uses significantly less computer memory than standard SVD based techniques.

### Acknowledgements

The authors would like to thank the organizers and funders of the CapoCaccia and Telluride Cognitive Neuromorphic Engineering Workshops, at which the bulk of this work was performed.